# A Projection Algorithm for the Unitary Weights

Hao-Yuan Chang
Electrical & Computer Engineering
University of California, Los Angeles

## Abstract

Unitary neural networks are promising alternatives for solving the exploding and vanishing activation/gradient problem without the need for explicit normalization that reduces the inference speed. However, they often require longer training time due to the additional unitary constraints on their weight matrices. Here we show a novel algorithm using a backpropagation technique with Lie algebra for computing approximated unitary weights from their pre-trained, non-unitary counterparts. The unitary networks initialized with these approximations can reach the desired accuracies much faster, mitigating their training time penalties while maintaining inference speedups. Our approach will be instrumental in the adaptation of unitary networks, especially for those neural architectures where pre-trained weights are freely available.

## 1. Introduction to unitary neural networks

We provide a brief introduction to unitary neural networks in deep learning, an idea that is becoming increasingly important in state-of-the-art neural architectures, the problem statement that we attempt to address, and our contribution in solving the stated problem.

### 1.1 Prior art

Unitary neural nets have existed in the deep learning community for many years. It is often disguised under different names such as orthogonal weights (Wang et al., 2020), orthogonality constraints (Edelman et al., 1998), or unitary recurrent neural networks (Lezcano-Casado & Martínez-Rubio, 2019). However, the central idea is the same. We try to force a mathematical constraint on the neural nets' weights to make them more stable and efficient. Unitary weights have a wide range of applications. For example, researchers have proposed to construct filters orthogonal to each other in the higher dimensional space. Such constrain results in unique filters; therefore, they are more efficient in recognizing different features (Huang et al., 2017). Most of the works focus on the advantages of unitary weights in maintaining the L2 norm of the neural signals. Typically, deep neural networks have many layers, and signals can grow or die depending on each layer's weight matrices. More precisely, the eigenvalues of these weight matrices. Even for recurrent neural nets, a single weight matrix expands out to multiple layers in time. Without normalizations, deep neural networks are unstable.

When the activations exceed the maximum representable value by a floating-point number, not a number error occurs. Moreover, as an example, if we assume each layer has a weight matrice with an eigenvalue of 10, after 50 hidden layers, the neural signal at the input of the network would be amplified $10^{50}$ times. The maximum representative number in IEEE single-precision floating-point format is around $10^{38}$. Single-precision float is the most popular floating-point format supported in graphics processing units and tensor-based hardware accelerators. Any number larger than $10^{50}$ will result in a not a number error and makes the backpropagation of training signals near this operating region unstable.

On the contrary, if each layer has an eigenvalue of 0.1, the neural activations will attenuate below the smallest representable floating-point number, $10^{-38}$, in the single-precision float representation. Training is slow when the neural activations are close to zero because the small backpropagated neural signals cause microscopic updates to the weight matrices. One way to make them stable is to employ unitary constraints on the weights. With the unitary weights, we can avoid the vanishing and exploding gradient problem.

### 1.2 Problem statement

There are many conventional normalization strategies such as batch norm (Ioffe & Szegedy, 2015), group norm (Wu & He, 2018), layer norm (Ba et al., 2016), and online norm (Chiley et al., 2019). These normalizations reset the neural signals to unit norm forcefully after each layer, regardless of how much the weights have amplified or reduce the norm. They often slow down inference if running statistics are required, and therefore, the normalization layers cannot be absorbed. Furthermore, the accuracy now depends on

———

the dataset statistics. By forcing the weight to be unitary during training, we eliminate the need for normalizations and speed up the inference time.

However, the problem with the existing framework using unitary weights is that the training time is often increased significantly. Most unitary networks use Lie algebra (Gilmore & Hermann, 1974) to guarantee that the weight matrices are unitary or orthogonal. Lie algebra is a technique to convert a manifold to a group representation through matrix exponentiation. We visualize this process's geometrical interpretation as the following: Lie parameterization is an infinitesimal direction in a manifold, and the exponentiation procedure will expand this small directional vector into a complete rotation in the n-dimensional space. Matrix exponentiation is an active field of study in mathematics with many recent improvements. Nevertheless, even with the best exponentiation algorithm, speed is still often an issue. Consequently, our work focuses on providing an algorithm that can approximate the unitary weights from networks pre-trained with non-unitary weights and explicit normalizations.

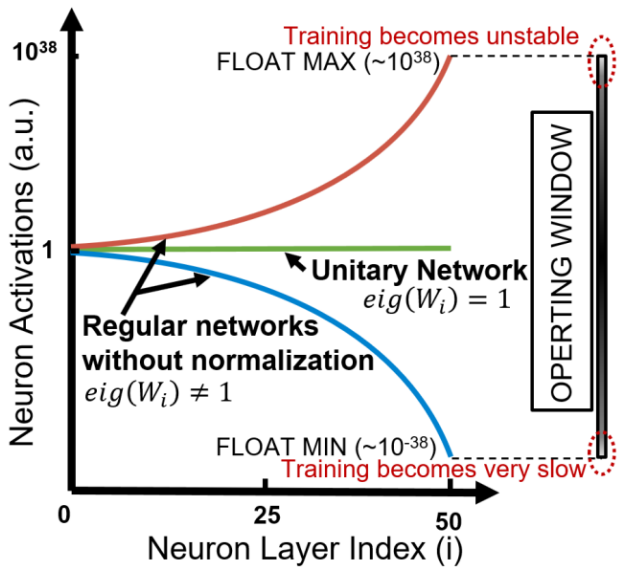

*Figure 1*. The problem of non-unitary weight matrices. Deep neural networks, by default, do not have any constraints on their synaptic weights, $W_i$, where i is the index of the hidden layer. Non-unitary $W_i$ with eigenvalues larger or smaller than one with magnifies or attenuate the neural activations across hidden layers, as illustrated in the figure's red and blue curves, respectively.

### 1.3 Our contribution

The deep learning community has moved towards an ecosystem where developers share pre-trained weights for most modern neural network architectures. As a result, we believe it is crucial to develop a new framework to take an existing pre-trained network and convert it to the unitary version. To achieve this aim, we record the regular networks' activations and find the closest equivalent unitary weights to deliver the same input/output characteristics. By effectively projecting the activations back to the unit sphere through a parallelizable, layer-wise backpropagation technique, we efficiently compute the equivalent unitary weights for zero-shot learning.

## 2. The projection algorithm

Assuming we have a set of weights that have been trained on a neural net with normalization blocks enabled, we can record each layer's activations and find the closes equivalent unitary weight implementation. Unitary matrices are constructed from Lie parameters using matrix exponentiation techniques (Humphreys, 1972); therefore, we can backpropagate the gradients back to the Lie parameters and find a set of Lie parameters result in behaviors closest to the one with normalization enabled.

In the following algorithm, $a_i$ is the activation of the input to a particular layer, and $a_i$ is the activation at the output of the layer before nonlinearity. Depth d is the total number of layers in the neural network. Lie parameters $L$ is a lower triangular traceless matrix. In regular neural networks, non-unitary weight matrices amplify or damp the neural signals, relying on normalization blocks to scale the signal back to the unit norm. This scaling process is a nonlinear operation and depends on the input activation data. Therefore, it is impossible to correctly translate a set of neural weights with normalization into a unitary version. Only an approximation is possible.

The problem we are trying to solve is essentially a least-square linear regression fitting problem. Our unitary projection algorithm is a gradient descent-based method to find the best fit for this projection. The theoretical worse case computational complexity for fitting this class of linear regression problem is proven to be O(n log m), where n is the number of data points and m is the degree of the model (Li, 1996), although the real run time is often much faster in practice.

---

**Algorithm 1** Unitary Projection

---

**Input**: activations $a_i$, depth $d$
**Output**: Lie parameters $L$
Initialize $L$ randomly
**repeat**
  **for** $i = 1$ to d
    $W = \exp(L - L^T)$
    $y\_predict = $ matrix_multiply($W, a_i$)
    $y\_true = a_{i+1}$
    $cost = $ mean_square_error($y\_predict, y\_true$)
    backpropagate *cost* and update $L$ with RMSprop
  **end for**
**until** *cost* stabilizes or reaches below a threshold

---

## 3. Experiment

We experiment using a 50-layer Fourier convolutional neural network to showcase zero-shot learning effectiveness using the proposed unitary projection algorithm.

### 3.1 Neural network architecture

To demonstrate the advantages of using unitary weight matrices, we devised the following neural network architecture and the MNIST dataset as an example. The inputs are a series of grayscale images with a dimension of 28 by 28 pixels. Because the input images have 28 x 28 grayscale pixels, each layer's activation map is a single matrix of size 28 x 28. Our framework's first step converts these images to the Fourier space by using the fast Fourier transformation (FFT). This step is necessary because we will be using matrix multiplication in the Fourier space as a convolution instead of the regular convolution in the spatial domain. The resulting transformation has both real and imaginary parts; hence, the neural activation map's dimensionality after the transformation will be 2 x 28 x 28. Because convolution in the space domain is matrix multiplication in the frequency domain, our convolution filter is a series of matrix multiplication between the neural activation map and the trainable unitary weight matrices. Our unitary network preserves the column-wise spectral power density of the activation signals across different neuron layers. The network consists of 50 fully connected neuron layers with dimensions identical to their inputs. *Figure 2* shows the network architecture for the MNIST-digit dataset. Each weight matrix is represented internally by the traceless lower triangular matrix. The final layer is for dimensionality reduction, which projects a vector of 1568 elements down to 10 via a 1568 x 10 fully connected layer. This final layer is not unitary. The output is a vector of size 10, representing the probability of being in 10 classes (i.e., from 0 to 9). The resulting 10-dimensional vector is then passed through a softmax function compared with the one-hot encoded ground-truth label.

An important detail to note is that the real and the imaginary parts are processed separately. That is the real part of the activation map (a 28 x 28 matrix) multiplied by a unitary matrix of the size 28 x 28, resulting in a matrix of the same size. Similarly but independently, the imaginary part is multiplied by a different trainable unitary weight matrix. The resulting activation map will pass through an element-wise tanh nonlinearity. This process is repeated 50 times for implementing a deep neural network with each layer extracting various features of the image. Before the final output, we collapse the neural activation maps (both real and imaginary parts) down to a vector of size 10, representing the probability of being in each class.

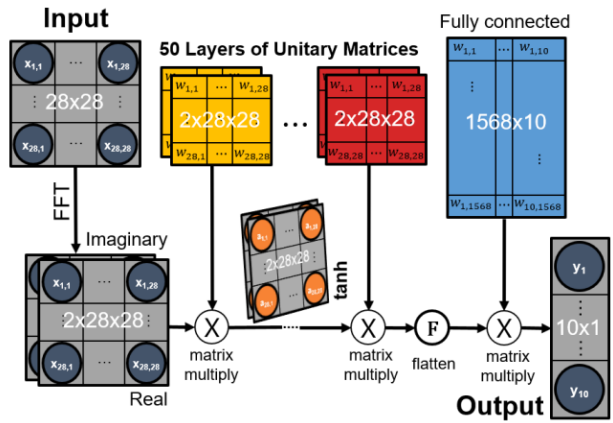

*Figure 2.* Unitary Convolutional Neural Network. We illustrate the architectural details of the unitary neural network designed to recognize hand-written digits. This illustration shows a batch size of one for clarity; in practice, we used a batch size of 512 to speed up the training process.

### 3.2 Dataset

We compared the convolutional neural networks' accuracy using the Modified National Institute of Standards and Technology hand-written digits dataset (MNIST) has ten classes and 60,000 28 x 28 grayscale images. We further separate the dataset into training and validation with a 5:1 split.

## 4. Results

We divided our results into two parts—the first part confirms that our unitary network without normalization is functional. The second portion showcases our projection algorithm's capabilities in zero-shot learning and training speed enhancement.

The first set of experiments are designed to show that our unitary network is working. The unitary network can recognize hand-written digits up to 96% for unseen data, and the activation norm stays consistent throughout the network (*Figure 3* & *Figure 4*). When we impose unitary constraints on the weight matrices using Lie algebra, we observe significant object recognition accuracy improvements in the test sets. Unitary weights prevent the neural activations from amplifying or diminishing across layers in a deep neural network. On the other hand, the non-unitary neural networks suffer convergence instability during training because of the exploding/vanishing gradient problem, resulting in low accuracy. Consequently, we demonstrated that unitarity ensures the neural signals stay within bounds and enhances neural network trainability.

Also, our results validated the capability of our network on zero-shot learning with the unitary projection algorithm. We compare the training and inference accuracy and loss when we initialize the network with the weights we computed using our projection algorithm vs. the weights initialized using the popular Xavier initialization (Glorot & Bengio, 2010). Our method delivers a much better performance without training in accuracy and loss, proving that the unitary weights found using our projection algorithm are a good approximation to a neural layer with normalization (*Figure 5*).

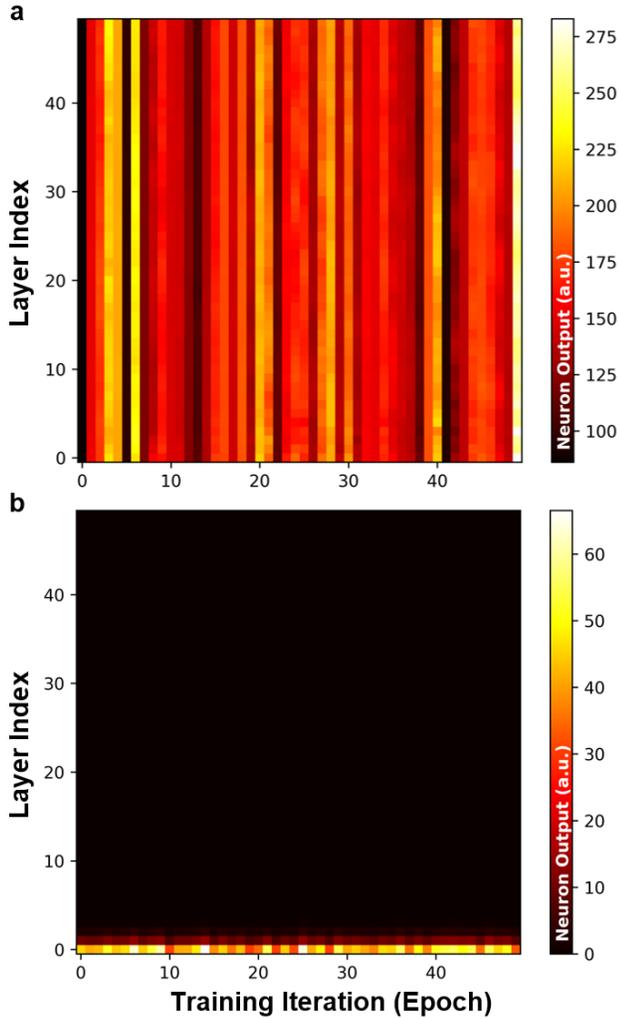

*Figure 3.* Neural activations (a) with unitary weights (b) with non-unitary weights. We plot the average neural activation per layer during network training for the MNIST dataset. The unitary network preserves the signal strength as it moves across neural layers. On the other hand, the conventional network has a damping effect on the neural signals.

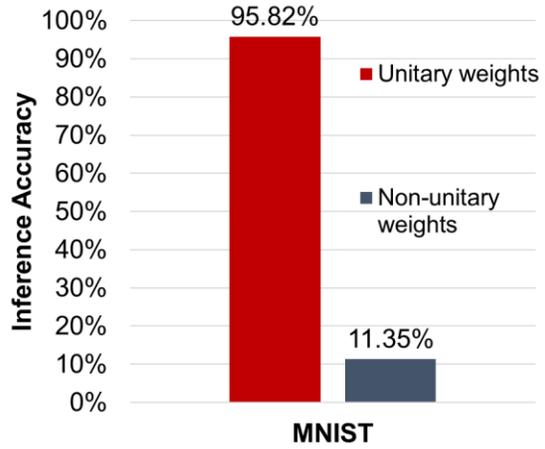

*Figure 4.* Accuracy of the convolutional neural network with and without unitary weights. Our unitary network delivers better performance comparing to the one without any normalization methods.

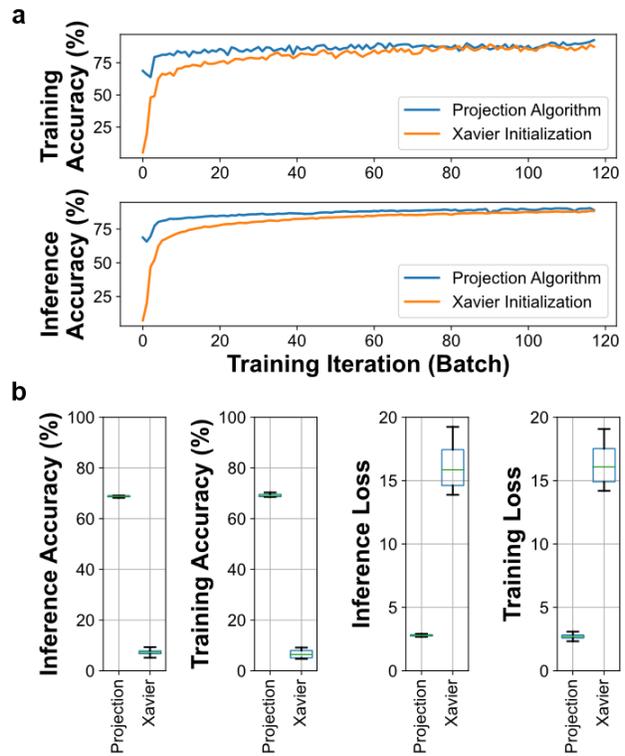

*Figure 5.* Zero-shot learning performance comparison a) during the first training epoch b) without any training. We compare the unitary network's performance initialized with weights computed from our prosed projection algorithm and Xavier initialization. These are box charts showing the overall statistics over four simulations.

## 5. Conclusion

This work demonstrates significant improvements in zero-shot learning accuracy and few-shot learning training speed with a unitary neural network initialized with our unitary projection algorithm. Our algorithm converts a network with trained weights to their closest unitary equivalents. By measuring the improvement on zero-shot performance, we confirm that our algorithm can find the closest approximate, reducing training time for the unitary neural networks.

# Supplementary Material

## 1. The Projection Algorithm

We assume that a set of standard, pre-trained weights are available. Such weights can be trained locally or downloaded as part of any published neural network model. We further assume it is possible to record the neural net's activation maps with these pre-trained weights. Source code is available as a part of the supplementary material.

## 2. Dataset

We used the standard Modified National Institute of Standards and Technology (MNIST) dataset that contains 60,000 images of hand-written digits. Each image has 28 pixels by 28 pixels. We downloaded the dataset with pre-separated training and validation sets from http://yann.lecun.com/exdb/mnist/. The training set has 50,000 images, and the validation set has 10,000 images. No data were excluded. However, we preprocess each image is the fast Fourier transform algorithm to construct a 2x28x28 tensor. That is the real and imaginary parts of the frequency domain signal.

## 3. Experiment Details

This section documents the experimental setups we used for data reproducibility.

### 3.1 Hyper-parameters

We summarize the hyper-parameters we used with the following tables.

| Parameter | Value |
| --- | --- |
| Learning Rate | 0.0001 |
| Initialization | Our projection algorithm or Xaviar initialization |
| Activation Function | tanh |
| Dropout | None |
| Optimizer | RMSprop |
| Batch Size | 512 |
| Number of Training Epochs | 100 for Figure 4<br>1 for Figure 5 |
| Number of Layers | 51 layers |
| Number of Neurons per Layer | 2x28x28 |
| Loss | Cross entropy loss |

*Table 1.* The hyper-parameters use in our neural network.

| Parameter | Value |
| --- | --- |
| Learning Rate | 0.0001 |
| Optimizer | RMSprop |
| Batch Size | 512 |
| Number of Epochs | 10 |
| Loss | Mean square error |

*Table 2.* The hyper-parameters use in our projection algorithm.

### 3.2 Experiment Setups

We run four training and four evaluations for the comparison between our projection algorithm and the Xavier initialization. There is no discrepancy between our result with the Xavier initialization with previously reported test accuracies. The terminal inference accuracy reaches above 95%, as expected for a deep neural network with similar architectures. Moreover, our experimental procedure is typical in this field. In Figure 4, training and evaluation take about 2560 seconds. In Figure 5, our pipeline has three components: first, we train the non-unitary network with hyperparameters in Table 1 to sample the activation maps for the first 30,000 training images. This will give us a set of trained weights with desired input-output characteristics per layer. This step takes about 96.1 seconds. Second, we use our proposed projection algorithm (Algorithm 1) with the hyperparameters in Table 2 to find the closet unitary weights that match the desired input-output characteristics. The projection algorithm converges in 9.8 seconds. Lastly, we measure the converted unitary weights' performance. The evaluation step takes 256.4 seconds. We ran all our simulations in a single Nvidia RTX 3090 with 24GB of VRAM.

### 3.3 Definition of Performance Metrics Our contribution

There are five quantities that we measure in our work. The neuron output in Figure 3 reports the average L2 norm of the activation map for each layer. It measures the activation strength of the neurons to ensure signal stability. In Figure 4 and Figure 5, inference accuracy measures the percentage of time that the network predicts the correct label for images in the validation dataset. Training accuracy measures the percentage of time the network predicts the correct label for images in the training dataset. Inference loss measures the cross-entropy between the predicted label and the correct label for the training dataset. This metric is calculated per batch and averaged over all batches. Similarly, the same definition applies to inference loss for the validation dataset. The box chart in Figure 5 reports the statistics of four simulation runs.